\documentclass[letterpaper, 10 pt, conference]{ieeeconf}  
\usepackage{xcolor}

\IEEEoverridecommandlockouts                              

\usepackage{caption}
\usepackage{graphicx}
\usepackage{graphics} 
\usepackage{epsfig} 
\usepackage{mathptmx} 
\usepackage{times} 
\usepackage{amsmath} 
\usepackage{amssymb}  
\usepackage{bm}

\usepackage{xspace}
\usepackage{balance}
\usepackage{comment}
\usepackage{subfigure}

\usepackage{booktabs}

\usepackage{color, colortbl}
\definecolor{lightgray}{gray}{0.85}

\usepackage{xcolor}

\DeclareMathAlphabet{\mathcal}{OMS}{cmsy}{m}{n}

\title{\LARGE \bf
Ellipse Loss for Scene-Compliant Motion Prediction
}

\author{ 
  Henggang Cui$^{1}$$^{*}$, Hoda Shajari$^{2}$$^{*}$, Sai Yalamanchi$^{1}$, Nemanja Djuric$^{1}$ \\
  \thanks{
    $^{1}$Uber Advanced Technologies Group (ATG), 50 33rd Street, Pittsburgh, PA 15201; emails:
     {\tt\small \{hcui2, syalamanchi, ndjuric\}@uber.com}
  }
  \thanks{
    $^{2}$University of Florida, work done during internship at Uber ATG; email: {\tt\small shajaris@ufl.edu}
  }
  \thanks{
    $^{*}$Authors contributed equally.
  }
}

\begin{document}

\maketitle
\thispagestyle{empty}
\pagestyle{empty}

\begin{abstract}
Motion prediction is a critical part of self-driving technology, responsible for inferring future behavior of traffic actors in autonomous vehicle's surroundings.
In order to ensure safe and efficient operations, prediction models need to output accurate trajectories that obey the map constraints.
In this paper, we address this task and propose a novel \emph{ellipse loss} that allows the models to better reason about scene compliance and predict more realistic trajectories.
Ellipse loss penalizes off-road predictions directly in a supervised manner, by projecting the output trajectories into the top-down map frame using a differentiable trajectory rasterizer module.
Moreover, it takes into account actor dimensions and orientation, providing more direct training signals to the model.
We applied ellipse loss to a recently proposed state-of-the-art joint detection-prediction model to showcase its benefits.
Evaluation on large-scale autonomous driving data strongly indicates that the method allows for more accurate and more realistic trajectory predictions.

\end{abstract}

\section{Introduction}

An autonomous vehicle is a highly involved system, supported by several input sensors (such as LiDARs, radars, and cameras), required to handle a large number of complicated situations in an uncertain environment \cite{Bertha2015}.
Following the completion of the perception step where the current state of the world is inferred, one of the most critical parts of the self-driving vehicle (SDV) framework is predicting how will the world state then evolve in the near future, a task addressed by the motion prediction module of the SDV.
Understanding the motion of surrounding actors, as well as the uncertainty of that motion, is necessary for ensuring safe and efficient routing of the SDV through a stochastic world.
Despite its importance, the task of motion prediction is still far from being solved, and significant research efforts are being invested in furthering the current state-of-the-art.

An important aspect of the motion prediction problem is ensuring that the inferred trajectories are \emph{scene compliant} and that they obey the constraints imposed by the underlying map~\cite{ridel2019scene, jain2019discrete, wang2020improving}.
This is however not a simple task.
The map data presents highly structured information, and it is not immediately clear how to utilize it in the learning model to help constrain the outputs and improve scene compliance.
Recent work proposed to provide it as an extra input \cite{luo2018fast,djuric2020multixnet}, to incorporate it as a part of the loss \cite{wang2020improving}, or as a post-processing step \cite{yalamanchi2020itsc}.
Nevertheless, the existing methods commonly struggle with this requirement, and more work is required to improve the scene compliance performance of the state-of-the-art motion prediction models. 

We focus on this important problem and propose a method to enhance scene compliance of predicted trajectories.
We propose a \emph{Box-aware Differentiable Trajectory Rasterizer} (BDTR) module to project the output trajectories into the top-down map, extending ideas proposed in \cite{wang2020improving}.
However, unlike the previous work that treats each actor as a point~\cite{wang2020improving}, BDTR explicitly incorporates actor's bounding box dimensions and orientation information.
We propose a novel \emph{ellipse loss} that employs BDTR to align the map and trajectory representations and penalize off-road trajectory predictions in a supervised manner to improve their scene-compliance.

We summarize the contributions of our work below:
\begin{itemize}
\item we propose the BDTR module that projects output trajectories into a top-down map while incorporating actor's bounding box and orientation, extending beyond point processes used by the current state-of-the-art;
\item we propose an ellipse loss that uses BDTR in a supervised manner to directly penalize off-road predictions;
\item the proposed ellipse loss is general, and we demonstrate its effectiveness by applying it to a state-of-the-art joint detection-prediction model, resulting in improved prediction performance on a large-scale real-world autonomous driving data set.
\end{itemize}

\section{Related work}

\subsection{Trajectory prediction for autonomous driving}

Trajectory prediction is one of the most critical tasks in an autonomous system, allowing SDVs to safely navigate in a uncertain real-world environment.
Given the historical observations, the trajectory prediction module predicts the future positions of the given actors, which can then be consumed by the motion planning module of the SDV.
The current state-of-the-art approaches rely on the deep neural network models for this task.
They commonly use an encoder network to generate embeddings from the historical observations (e.g., by using recurrent layers) and a decoder network to generate future trajectories as well as their uncertainties (e.g.,~\cite{lee2017desire, ivanovic2019trajectron, salzmann2020trajectron++, chai2019multipath, gupta2018social, zhao2019multi, sadeghian2019sophie, kosaraju2019social}).
In the cases when there is map information available, a common approach is to rasterize the map polygons (e.g., lanes and other drivable surfaces) into a multi-channel bird's-eye view (BEV) image, and use a convolutional neural network to extract scene context features from the BEV image which can then be fused with the embeddings pertaining to the historical observation \cite{luo2018fast, djuric2020multixnet,salzmann2020trajectron++, chai2019multipath, cui2019multimodal, djuric2020uncertainty, chou2020iv}.

The \emph{waypoint representation} is the most popular trajectory representation used by most state-of-the-art prediction models (e.g.,~\cite{salzmann2020trajectron++, chai2019multipath, cui2019multimodal, djuric2020uncertainty, chou2020iv}),
where each trajectory is represented as a sequence of future position coordinates $\{(x_t, y_t)\}_{t=1}^{T_h}$, where $T_h$ is the prediction horizon.
This representation is compact and allows for an efficient interface between the prediction and the motion planning modules of the autonomous driving system.
Some researchers proposed to use an \emph{occupancy grid representation} that encodes the future motion prediction as an occupancy grid \cite{jain2019discrete, bansal2018chauffeurnet, ridel2019scene}.
However, while the trajectory representation can be exactly cast into an occupancy grid the opposite is not true, leading to the loss of detailed information that some online systems rely on. 
Moreover, the grid representation is less compact, resulting in larger memory and latency costs which makes this approach less commonly used in applied systems.

The state-of-the-art prediction models are often trained in a supervised fashion, using some sort of distance measure (e.g., L2, smooth L1, KL divergence, or log-likelihood) between the output trajectories and the ground-truth trajectory as their loss function (e.g.,~\cite{djuric2020multixnet,ivanovic2019trajectron, salzmann2020trajectron++, chai2019multipath, cui2019multimodal, djuric2020uncertainty, chou2020iv}).
In recent years Generative Adversarial Networks (GANs) have shown to reach good performance \cite{wang2020improving,gupta2018social, sadeghian2019sophie, kosaraju2019social}, where the trajectory decoder is trained with the help of a discriminator network.
However, GAN-based models are rarely used in real-world autonomous systems as they require one to draw many samples to sufficiently cover the trajectory space (e.g., as many as $20$ samples in~\cite{wang2020improving, gupta2018social, sadeghian2019sophie}), which makes them impractical for time-sensitive applications.

When it comes to model inputs, some works assume access to a separate detection and tracking module that infers historical actor tracks ingested by the model (e.g.,~\cite{lee2017desire, salzmann2020trajectron++, kosaraju2019social, cui2019multimodal}).
On the other hand, recently proposed approaches directly take in the raw LiDAR point clouds as their inputs and jointly perform object detection and trajectory prediction in the same network (e.g.,~\cite{luo2018fast,djuric2020multixnet, casas2018intentnet, zeng2019end, casas2019spatially, wu2020motionnet, meyer2020laserflow}).
The benefit of such joint models is that there is no loss of information between the detection and prediction modules, where the prediction module has access to richer information from the raw sensor data.
Our work focuses on these models and proposes a novel loss that can be directly applied to improve prediction performance.

\subsection{Scene-compliant trajectory prediction}

Traditional loss functions such as L2, smooth L1, or KL divergence used in previous works focus on minimizing the distance from the prediction to the ground truth, however, they do not have any explicit mechanism to enforce the trajectories to be scene-compliant (e.g., stay within the drivable region).
In order to encourage the predicted trajectories to be scene-compliant, some works have proposed various types of scene-compliant loss functions to explicitly penalize off-road predictions, which we discuss in this section.

ChauffeurNet~\cite{bansal2018chauffeurnet}, DRF-NET~\cite{jain2019discrete}, and Ridel et al.~\cite{ridel2019scene} proposed to represent each predicted trajectory waypoint as an occupancy grid and computed the product of the occupancy grid and a non-drivable region mask as one of their loss terms to penalize off-road predictions.
To obtain trajectories in the form of waypoint representation, as a post-processing step Ridel et al.~\cite{ridel2019scene} sampled the waypoint positions from each predicted occupancy grid.
This approach, however, cannot guarantee that the waypoints sampled from a scene-compliant occupancy grid will remain scene-compliant.
Niedoba et al.~\cite{niedobaimproving} proposed a scene-compliant loss that can be directly applied to the waypoints.
In their work, they identified predicted waypoints that are not scene-compliant and upweighted their loss by a fixed factor, so the model focuses more on such cases.
Wang et al.~\cite{wang2020improving} proposed a \emph{Differentiable Trajectory Rasterizer} (DTR) module that bridges the waypoint and the occupancy grid representations.
DTR converts a waypoint coordinate $(x, y)$ into an occupancy grid in a differentiable manner, so the losses previously only applicable to the occupancy grids can be applied to the waypoint representations as well.
Different from our work, Wang et al.'s DTR module~\cite{wang2020improving} treated each actor as a point and focused on keeping the actor centroid inside the drivable region instead of the whole bounding box.
Moreover, Wang et al.~\cite{wang2020improving} used a GAN architecture and applied their DTR in a discriminator network to make it more sensitive to trajectories that are not scene-compliant.
On the other hand, we apply our box-aware DTR module in a supervised manner and compute our ellipse loss as the product of the occupancy grid and the non-drivable region mask, similarly to \cite{ridel2019scene,jain2019discrete,bansal2018chauffeurnet}, resulting in a more flexible and easier to use architecture.

Instead of treating each actor as a point as in previous works~\cite{ridel2019scene,jain2019discrete,wang2020improving,niedobaimproving}, we fully consider the actor bounding box dimensions by proposing a novel Box-aware DTR (BDTR) module.
The BDTR module has two important distinctions compared to DTR.
First, we use a 2D Gaussian distribution parameterized by an actor's dimensions, as opposed to a fixed distribution. 
Second, we truncate the rasterized occupancy grid based on the object's bounding box dimensions.
These extensions allow the proposed ellipse loss to make the entirety of the actor's bounding box scene-compliant instead of just the centroid.
The evaluation results indicate that the introduced extensions are critical in improving the prediction performance in a supervised setting.


\section{Ellipse Loss}
In this section we introduce our novel scene-compliant \emph{ellipse loss}.
While the proposed approach is generic and can be applied to any motion prediction model, we implemented and evaluated it on top of MultiXNet~\cite{djuric2020multixnet}, a state-of-the-art joint object detection and motion prediction model.

\subsection{Backbone network and output representation}

The MultiXNet model takes a total of $T_h$ current and historical LiDAR sweeps as input, as well as the high-definition map of SDV's surroundings (containing $M$ different map elements).
Assuming the area of interest centered on SDV to be of length $L$, width $W$, and height $V$, the method voxelizes the LiDAR points into a 3D tensor $\mathcal{V}$ of shape $L \times W \times (V T_h)$ with a voxel size of $\Delta_L \times \Delta_W \times \Delta_V$, and rasterizes the map onto a multi-channel binary image $\mathcal{M}$ of shape $L \times W \times M$ using the same resolution. 
Moreover, we assume that we are provided with a binary \emph{drivable region mask} $\mathcal{D}$ of the same dimensions and resolutions as the LiDAR and map rasters, with drivable cells encoded as $1$ and non-drivable cells as $0$.

The model uses a multi-scale convolutional network as the \emph{first-stage} network, which runs on the concatenation of $\mathcal{V}$ and $\mathcal{M}$ and produces the probability that each grid cell contains an object and the corresponding object bounding box parameters.
The detections (i.e., bounding boxes with high objectness probabilities) are then passed as input to the \emph{second-stage} network, which then outputs their future trajectory predictions (represented as a sequence of future waypoints) for a total of $T_f$ time steps.
The waypoint of the $i$-th object at each time step $t$ is parameterized as ${\bf s}_t^{(i)} = (x_t^{(i)}, y_t^{(i)}, l_t^{(i)}, w_t^{(i)}, \theta_t^{(i)})$, representing its center position, box length, box width, and box orientation, respectively.
We assume that the bounding box remains constant over time, and use the same box size $(l^{(i)}, w^{(i)})$ for all time steps $t$.

\subsection{Loss function}

The first-stage and second-stage networks are trained end-to-end.
A focal loss term is applied to the object detection score of each grid cell from the first-stage network.
Since we applied our ellipse loss to the trajectory predictions from the second-stage network, we will focus on discussing the second-stage prediction losses in the remainder of the paper.


The baseline MultiXNet model optimizes the following loss on the second stage outputs,
\begin{align}
\begin{split}
\label{eq:vanilla_loss}
\mathcal{L}_\text{vanilla} = \sum_{i=1}^N \sum_{t=1}^{T_f} &\ell_1 \big(x_t^{(i)} - x_t^{(i)*} \big) + \ell_1 \big(y_t^{(i)} - y_t^{(i)*} \big) \\
&+ \ell_1 \big(l_t^{(i)} - l_t^{(i)*} \big) + \ell_1 \big(w_t^{(i)} - w_t^{(i)*} \big) \\
&+ \ell_1 \big(\sin (\theta_t^{(i)}) - \sin (\theta_t^{(i)*}) \big) \\
&+ \ell_1 \big(\cos (\theta_t^{(i)}) - \cos (\theta_t^{(i)*}) \big),
\end{split}
\end{align}
where $N$ is the number of true positive detections, $\ell_1$ represents the smooth-$\ell_1$ loss, and $(x_t^{(i)*}, y_t^{(i)*}, l_t^{(i)*}, w_t^{(i)*}, \theta_t^{(i)*})$ are their ground-truth waypoints.

We extend MultiXNet by adding the proposed ellipse loss as an additional loss term to make trajectory predictions more scene-compliant\footnote{The ellipse loss only encourages the predictions to stay inside the drivable region, so the original prediction loss \eqref{eq:vanilla_loss} is still required to encourage the predicted trajectories to be close to the ground truth.}.
Given a predicted waypoint ${\bf s}_t^{(i)} = (x_t^{(i)}, y_t^{(i)}, l_t^{(i)}, w_t^{(i)}, \theta_t^{(i)})$, we use our novel Box-aware DTR module to convert it to a 2D occupancy grid $\mathcal{G}({\bf s}_t^{(i)})$  (referred to as \emph{Gaussian raster} and discussed in detail in the following section) with the same dimensions $L \times W$ and grid cell size  $\Delta_L \times \Delta_W$ as the drivable region mask $\mathcal{D}$.
Then, we compute the ellipse loss $\mathcal{L}_\text{ellipse}$ as follows,
\begin{equation}
\label{eq:ellipse_loss}
\mathcal{L}_\text{ellipse} = \sum_{i=1}^N \sum_{t=1}^{T_f}  I_t^{(i)*} \sum_\text{cells} \mathcal{G}({\boldsymbol s}_t^{(i)}) \circ (1 - \mathcal{D}),
\end{equation}
where $1 - \mathcal{D}$ is the non-drivable region mask, $\circ$ denotes the Hadamard product,
and $I_t^{(i)*}$ is an indicator term equal to $1$ when the ground-truth box ${\bf s}_t^{(i)*}$ is inside the drivable region and $0$ otherwise.
We included $I_t^{(i)*}$ in the loss to avoid penalizing off-road predictions when the ground truth is already off-road (e.g., parked cars on the side of the road).

Then, the final second-stage loss is computed as
\begin{equation}
\label{eq:total_loss}
\mathcal{L}_\text{ours} = \mathcal{L}_\text{vanilla} + \lambda \mathcal{L}_\text{ellipse},
\end{equation}
where the ellipse loss is weighted by a fixed parameter $\lambda$.
Note that BDTR is only used during training for the loss computation, and does not impact model inference latency.

\begin{figure}
    \centering
    \includegraphics[width=0.43\textwidth]{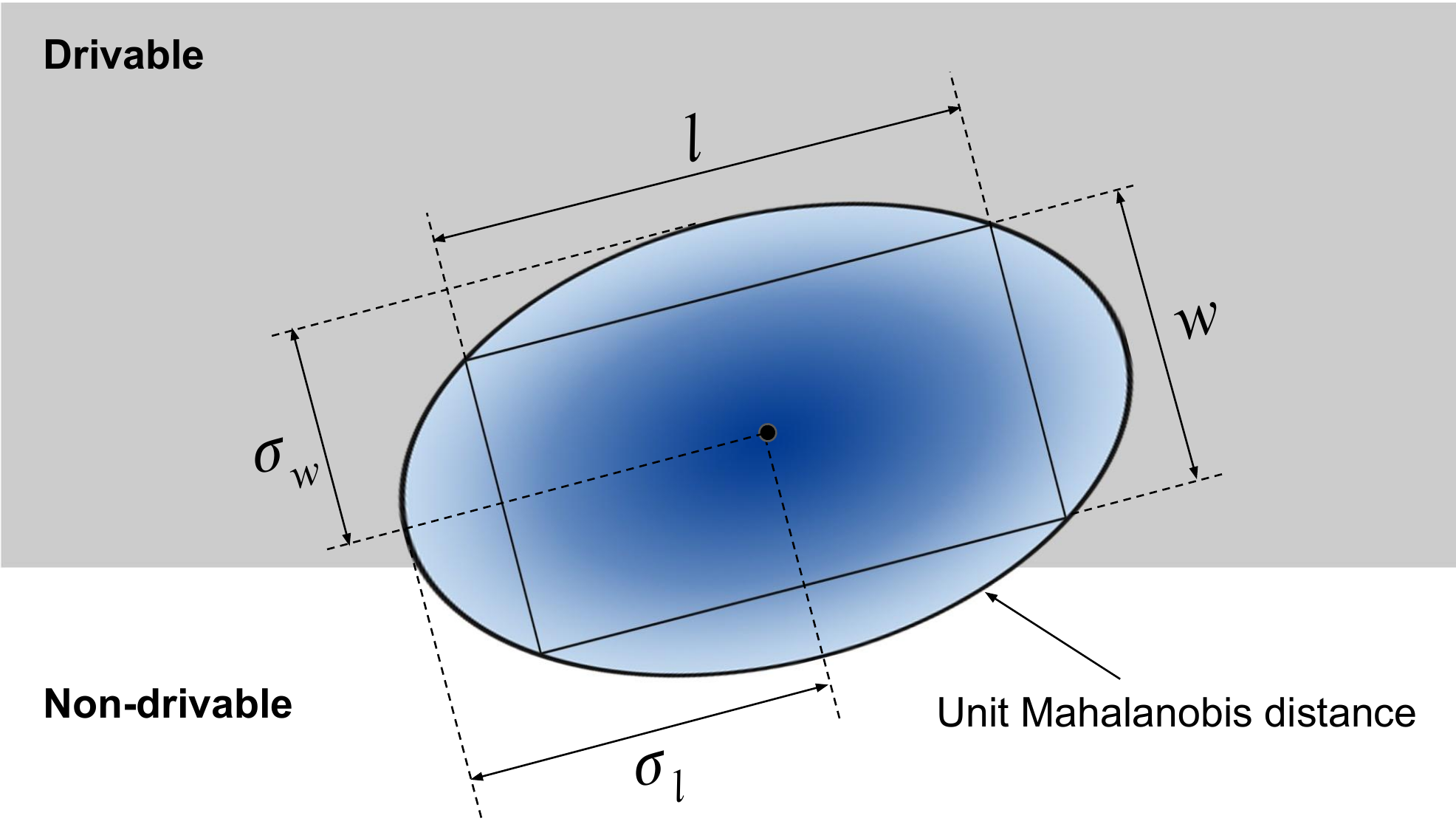}
    \caption{Illustration of the ellipse loss; inscribed rectangle denotes the actor's bounding box, blue ellipse is the unit Mahalanobis distance range in which the loss is computed}
    \label{fig:ellipse_loss}
    \vspace{-0.2cm}
\end{figure}

\begin{figure*}[h]
    \centering
    \subfigure{
        \includegraphics[width=2.5cm]{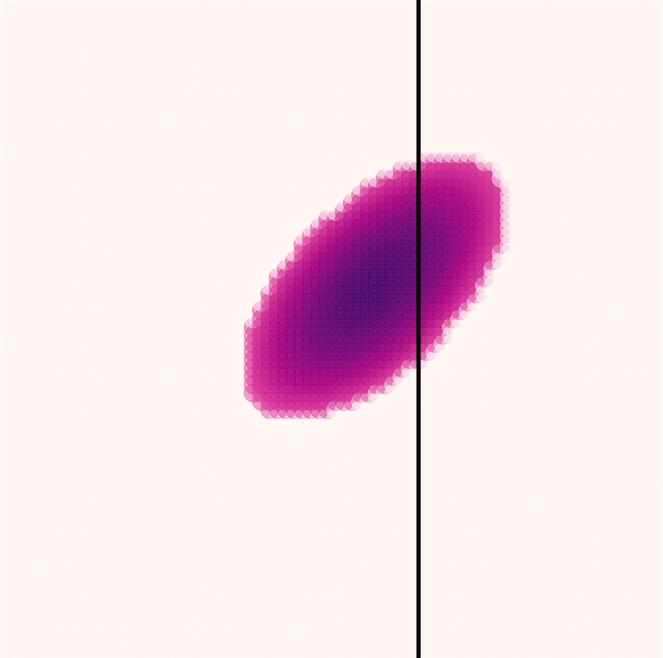}
    }
    \subfigure{
        \includegraphics[width=2.5cm]{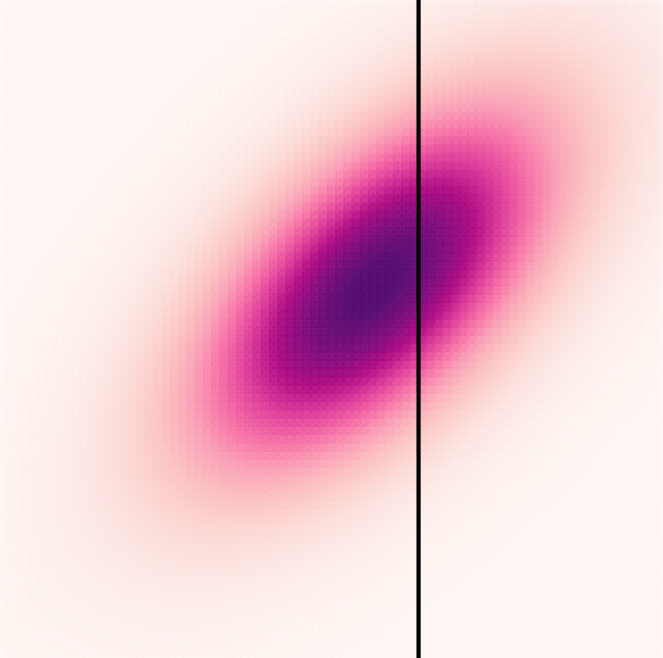}
        \label{fig:offroad-correction/no_truncation_before}
    }
    \hspace{1.0 cm}
    \subfigure{
        \includegraphics[width=2.5cm]{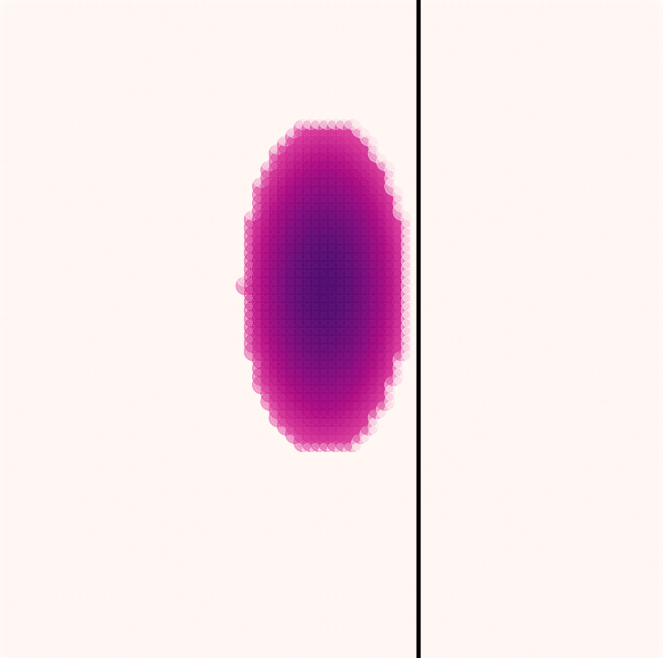}
    }
    \subfigure{
        \includegraphics[width=2.5cm]{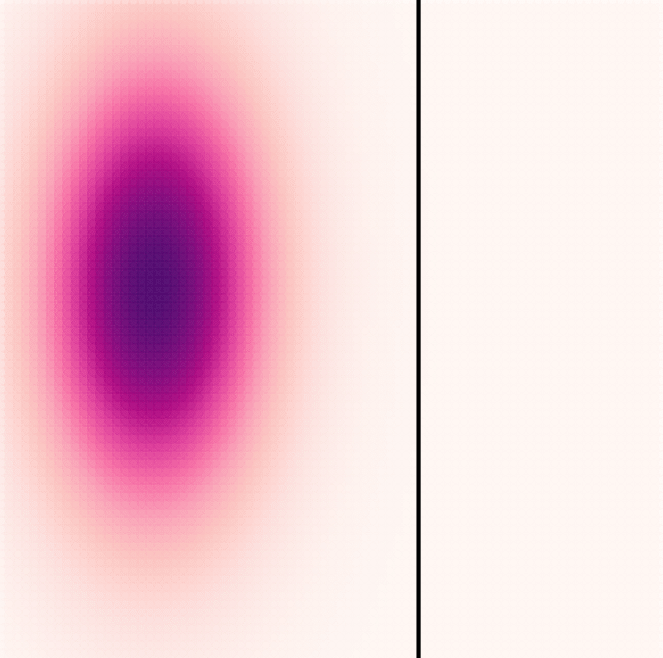}
    }
    \vspace{0.2 cm}
    \caption{Characterization of the ellipse loss on a toy example, showing that the loss pushes the actor away from the non-drivable region (right to the boundary line) by correcting both the center position and the orientation of the bounding box; the left two figures show the initial actor state, and the right two figures show the state after $1{,}000$ iterations of gradient updates for the ellipse loss with and without truncation, respectively}
    \label{fig:offroad-correction}
\end{figure*}

\subsection{Box-aware Differentiable Trajectory Rasterization}

In this section we present the novel Box-aware DTR module that converts a waypoint ${\bf s}_t^{(i)}$ into an occupancy grid $\mathcal{G}({\bf s}_t^{(i)})$.
Since the operation is applied independently to each actor and waypoint, for clarity we omit the actor index $i$ and time index $t$ in the remainder of the section.

BDTR computes a value of each cell $(i, j)$ in the grid $\mathcal{G}({\bf s})$ (denoted as $\mathcal{G}_{i, j}({\bf s})$) as the density of a 2D Gaussian $\mathcal{N}({\boldsymbol 0}, {\boldsymbol \Sigma})$ evaluated at ${\bf d}_{i, j}(x, y)$,
\begin{equation}
\label{eq:dtr}
\mathcal{G}_{i, j}(x, y, l, w, \theta) = \mathcal{N}\big({\bf d}_{i, j}(x, y); {\boldsymbol 0}, {\boldsymbol \Sigma}(l, w, \theta)\big).
\end{equation}
Value of ${\bf d}_{i, j}(x, y)$ is computed as a displacement vector from the waypoint center position $(x, y)$ to the grid cell $(i, j)$, 
where the grid cell position is in the same coordinate frame as the waypoint coordinate $(x, y)$.

{\bf Covariance matrix as a function of actor state.}
Different from the original DTR module~\cite{wang2020improving} where the covariance matrix is symmetric and hard-coded as $\boldsymbol \Sigma = \text{diag}(\sigma, \sigma)$ with a predefined $\sigma$,
we define ${\boldsymbol \Sigma}$ in \eqref{eq:dtr} as a function of the actor's predicted dimensions $l$ and $w$, as well as its predicted orientation $\theta$.
In particular, for each actor we set the standard deviation of each of the two axes of the Gaussian distribution as a linear function of the bounding box size $(l, w)$ as
\begin{equation}
\label{eq:decide_sigmas}
(\sigma_l, \sigma_w) = (k l, k w),
\end{equation}
where $k$ is a fixed positive scaling factor. We further rotate the covariance matrix to be aligned with the predicted actor's orientation $\theta$, computed as
\begin{equation}
\label{eq:rotation}
{\boldsymbol \Sigma}(l, w, \theta) = {\bf R}(\theta)^\top \text{diag}(\sigma_l, \sigma_w) {\bf R}(\theta),
\end{equation}
where ${\bf R}(\theta)$ represents a rotation matrix constructed from the actor's orientation $\theta$.



{\bf Truncation by bounding box dimension.}
Multiplying the entire Gaussian raster with the non-drivable region mask in the loss \eqref{eq:ellipse_loss} has a tendency to over-compensate and push the predictions too far away from the non-drivable region, which often results in worse performance, as illustrated in the following Section \ref{sec:gradient analysis} and empirically shown in Section~\ref{sec:ablation_study}.
To address this issue, we propose to truncate the Gaussian raster based on the actor's bounding box to limit its range. 

Earlier work has shown that the gradient magnitude of the Gaussian raster reaches its maximum at the unit Mahalanobis distance~\cite{wang2020improving}.
Based on this property, we propose to truncate the Gaussian raster at this distance, which corresponds to an ellipse with radii $\sigma_l$ and $\sigma_w$.
For each actor, we intend this ellipse to circumscribe the actor's bounding box $(l, w)$, to ensure scene-compliance while minimizing over-compensation.
As a result, we set $k = \frac{\sqrt{2}}{2}$ in \eqref{eq:decide_sigmas} and zero out all the values outside the truncation range.
We stop the gradients of $\mathcal{L}_\text{ellipse}$ w.r.t. $l$ and $w$ to prevent the model from minimizing the ellipse loss by incorrectly reducing the bounding box size.

Figure~\ref{fig:ellipse_loss} illustrates the range of the truncated Gaussian raster.
The intuition is that we only want to penalize predictions that cause the actor bounding box to be outside the drivable region, and when the bounding box is already outside the non-drivable region it should have zero loss.

\subsection{Characterization on a toy example}
\label{sec:gradient analysis}


We run simple toy experiments to exemplify how the proposed ellipse loss pushes away the non-scene-compliant actors from the non-drivable region, shown in Figure~\ref{fig:offroad-correction}.
In particular, we mark the region to the right of the black boundary line as non-drivable and place the actor's bounding box so as to straddle the boundary.
Two leftmost figures show the actor's initial position and orientation, as well as its Gaussian raster.
We then use the gradients from the ellipse loss to update the actor's position and orientation, running for $1{,}000$ iterations.
We performed two sets of experiments, one with the Gaussian raster truncated at unit Mahalanobis distance and another experiment with no truncation.

The final states after the gradient updates are plotted in the two rightmost figures.
Both the truncated and non-truncated ellipse losses were able to push the actor away from the non-drivable region by correcting both the center position and the orientation.
We can see that the truncated loss pushed the actor to be exactly at the boundary of the non-drivable region.
The non-truncated ellipse loss, on the other hand, kept pushing the actor further away from the boundary. 
Although it may seem that the loss resulting in a state that is further from the boundary might be a better choice, traffic actors in the real world do not behave in that way. 
This is also confirmed by the improved experimental performance of the truncated loss, presented later in Section~\ref{sec:ablation_study}.

\begin{table*} [t!]
\centering
\caption{Comparison of the competing methods;   confidence intervals are computed over 3 runs}
\label{tab:pred-errors}
{\normalsize
{
  \begin{tabular}{lcccccc}
     & \multicolumn{2}{c}{\bf ${\boldsymbol \ell_2}$ [m] $\downarrow$} & \multicolumn{2}{c}{\bf CtrORFP [\%] $\downarrow$} & \multicolumn{2}{c}{\bf BoxORFP [\%] $\downarrow$} \\
     \cmidrule(lr){2-3}\cmidrule(lr){4-5}\cmidrule(lr){6-7}
    {\bf Method} & {\bf Avg} & {\bf @3s} & {\bf Avg} & {\bf @3s} & {\bf Avg} & {\bf @3s} \\
    \hline
    \rowcolor{lightgray}
    \texttt{MultiXNet} \cite{djuric2020multixnet} & {\bf 0.465 {\footnotesize $\pm$ 0.001}} & {\bf 0.836 {\footnotesize $\pm$ 0.002}} & 0.061 {\footnotesize $\pm$ 0.007} & 0.094 {\footnotesize $\pm$ 0.018} & 0.624 {\footnotesize $\pm$ 0.009} & 0.850 {\footnotesize $\pm$ 0.029} \\
    \texttt{Off-Road}~\cite{niedobaimproving} & \bf \bf 0.465 {\footnotesize $\pm$ 0.002} & {\bf 0.837 {\footnotesize $\pm$ 0.005}} & 0.058 {\footnotesize $\pm$ 0.001} & 0.086 {\footnotesize $\pm$ 0.006} & 0.616 {\footnotesize $\pm$ 0.015} & 0.853 {\footnotesize $\pm$ 0.020} \\
    \rowcolor{lightgray}
    \texttt{Ellipse-Loss} & 0.472 {\footnotesize $\pm$ 0.004} & 0.846 {\footnotesize $\pm$ 0.008} & {\bf 0.031 {\footnotesize $\pm$ 0.003}} & {\bf 0.043 {\footnotesize $\pm$ 0.006}} & {\bf 0.445 {\footnotesize $\pm$ 0.007}} & {\bf 0.485 {\footnotesize $\pm$ 0.004}} \\
    \hline
\end{tabular}
}
\vspace{+0.4cm}
}

\end{table*}

\begin{figure*}[t!]
    \centering
    \subfigure{\includegraphics[width=0.38\textwidth]{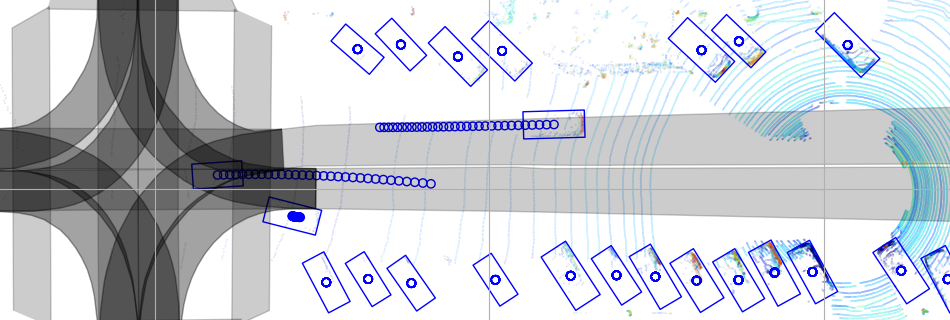}}
    \hspace{0.1 cm}
    \subfigure{\includegraphics[width=0.38\textwidth]{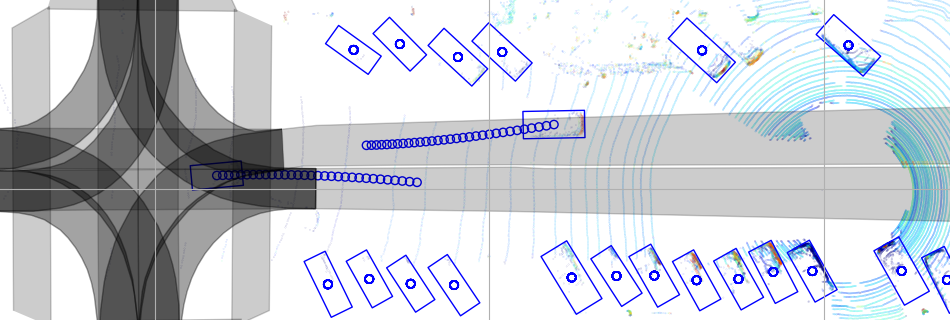}}\\
    \vspace{0.1 cm}
    \subfigure{\includegraphics[width=0.38\textwidth]{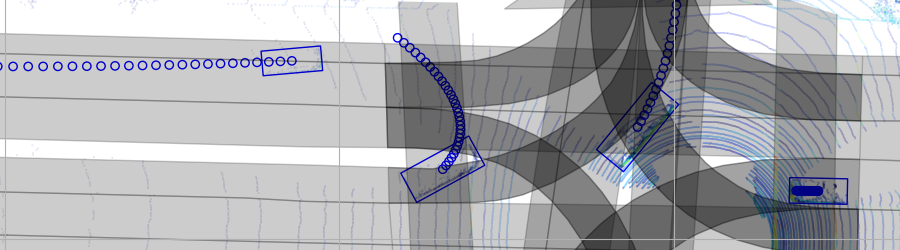}}
    \hspace{0.1 cm}
    \subfigure{\includegraphics[width=0.38\textwidth]{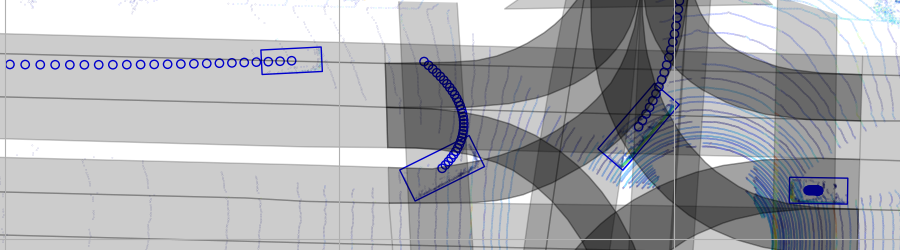}}\\
    \vspace{0.1 cm}
    \subfigure{\includegraphics[width=0.38\textwidth]{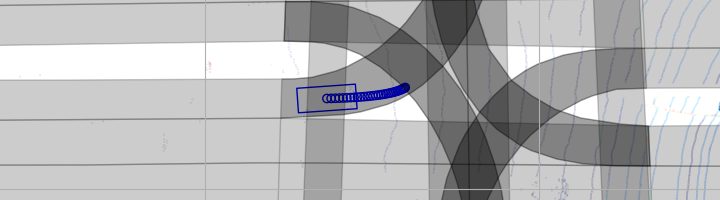}}
    \hspace{0.1 cm}
    \subfigure{\includegraphics[width=0.38\textwidth]{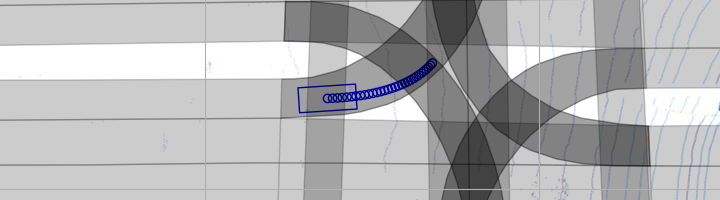}} 
    \caption{Case studies for \texttt{MultiXNet} (left) and \texttt{Ellipse-Loss} (right) on three representative scenarios; the detected objects and trajectory predictions are plotted in blue}
    \label{fig:qualitative_results}
\end{figure*}

\section{Evaluation}

\subsection{Experimental setup}

{\bf Data set.}
We evaluated the proposed ellipse loss on the ATG4D autonomous driving data set~\cite{djuric2020multixnet,meyer2019lasernet}.
The data was collected in several North American cities by a fleet of SDVs, each equipped with a 64-beam roof-mounted LiDAR, containing over $1$ million frames captured at $10Hz$ from $5{,}500$ different scenarios.
It also includes manually labeled object bounding boxes and detailed map polygons.
Since the data set does not have an explicit drivable region flag, we treated the union of all map components (including lanes, intersections, etc.) as the drivable region.
Note that an exact definition of the drivable region is orthogonal to our method, and one can also customize the drivable region for each individual actor to include only the lanes that the actor has access to, if such information is available in the data.

{\bf Implementation details.}
We used the same model architecture as \mbox{MultiXNet~\cite{djuric2020multixnet}} for all of our experiments,
with the only difference among the competing models being their loss functions.
We set the BEV voxel, map raster, and drivable region mask input shape and resolution to $L = 150\text{m}$, $W = 100\text{m}$, $V = 3.2\text{m}$, $\Delta_L = 0.16\text{m}$, $\Delta_W = 0.16\text{m}$, $\Delta_V = 0.2\text{m}$.
We used 1s of historical LiDAR inputs ($T_h = 10$ at $10\text{Hz}$) to predict 3s-long future trajectories ($T_f = 30$ at $10\text{Hz}$).
For the models employing the ellipse loss we set $\lambda = 0.03$, obtained through cross-validation.
The models were implemented in PyTorch~\cite{paszke2019pytorch} and trained using the same setup as MultiXNet~\cite{djuric2020multixnet}.

{\bf Evaluation metrics.}
While the method is general and can be applied to any actor type, in this work we focus on evaluating vehicle trajectory prediction.
For each competing model, we evaluated ${\ell_2}$ error of its trajectory predictions, as well as the \emph{off-road false positive} (ORFP) metrics \cite{wang2020improving} for both the final 3-second waypoint and averaged over the whole trajectory for each vehicle actor.
We refer to an output trajectory waypoint as being off-road false positive if it is predicted to be off-road while the corresponding ground-truth waypoint is in-road.
The ORFP ratio at a particular horizon is computed as the number of predictions that are ORFP at this horizon over the total number of predictions\footnote{If a predicted waypoint or a ground truth is outside the $150\text{m} \times 100\text{m}$ raster range, we re-use the ORFP result of a previous waypoint.}.
We used two policies to decide whether a waypoint is off-road or not.
The \emph{center-off-road} policy labels a waypoint off-road if the center $(x, y)$ position of the bounding box is outside the drivable region,
and the \emph{box-off-road} policy labels a waypoint off-road if any of the four corners of the bounding box is outside the drivable region.
We refer to these two metrics as CtrORFP and BoxORFP, respectively.

The competing models perform joint object detection and trajectory prediction, achieving nearly identical detection performance (result not shown).
In order to make sure the prediction metrics are fairly compared, we set the detection score threshold for each model such that they all achieve the same $0.8$ recall with an IoU threshold of $0.5$ for vehicles, as suggested by earlier work \cite{djuric2020multixnet,casas2019spatially}. 
Then, the prediction ${\ell_2}$ error and ORFP metrics are evaluated on true-positive detections computed at this threshold.

{\bf Baselines.}
We compared our approach against recently proposed state-of-the-art methods:
a \texttt{MultiXNet} baseline \cite{djuric2020multixnet} that uses the $\mathcal{L}_\text{vanilla}$ term from \eqref{eq:vanilla_loss} in the loss,
and \texttt{Off-Road} method~\cite{niedobaimproving} that directly penalizes the positions of off-road false positive waypoints.
In particular, following the setup from~\cite{niedobaimproving} we upweight the $\ell_1 \big(x_t^{(i)} - x_t^{(i)*} \big)$ and $\ell_1 \big(y_t^{(i)} - y_t^{(i)*} \big)$ terms of $\mathcal{L}_\text{vanilla}$ by a factor of $5$.

\begin{table*} [t!]
\centering
\caption{Ablation study of the ellipse loss with different truncation settings}
\label{tab:ablation_study_ellipse_sizes}
{\normalsize
{
  \begin{tabular}{lcccccc}
     & \multicolumn{2}{c}{\bf ${\boldsymbol \ell_2}$ @3s [m] $\downarrow$} & \multicolumn{2}{c}{\bf CtrORFP @3s [\%] $\downarrow$} & \multicolumn{2}{c}{\bf BoxORFP @3s [\%] $\downarrow$} \\
     \cmidrule(lr){2-3}\cmidrule(lr){4-5}\cmidrule(lr){6-7}
    {\bf Method} & {\bf Val.} & {\bf Rel.} & {\bf Val.} & {\bf Rel.} & {\bf Val.} & {\bf Rel.} \\
    \hline
    \rowcolor{lightgray}
    \texttt{0.5Md-ellipse} & 0.845 & -0.1\% & 0.056 & +30\% & 0.738 & +52\% \\
    \texttt{1Md-ellipse} & 0.846 & 0\% & 0.043 & 0\% & 0.485 & 0\% \\
    \rowcolor{lightgray}
    \texttt{2Md-ellipse} & 0.899 & +6\% & 0.033 & -23\% & 0.348 & -28\% \\
    \texttt{No-truncation} & 1.719 & +103\% & 0.017 & -60\% & 0.139 & -71\% \\
    \hline
\end{tabular}
}
}
\end{table*}


\begin{figure*}[t!]
    \centering
    \subfigure{\includegraphics[width=0.49\textwidth]{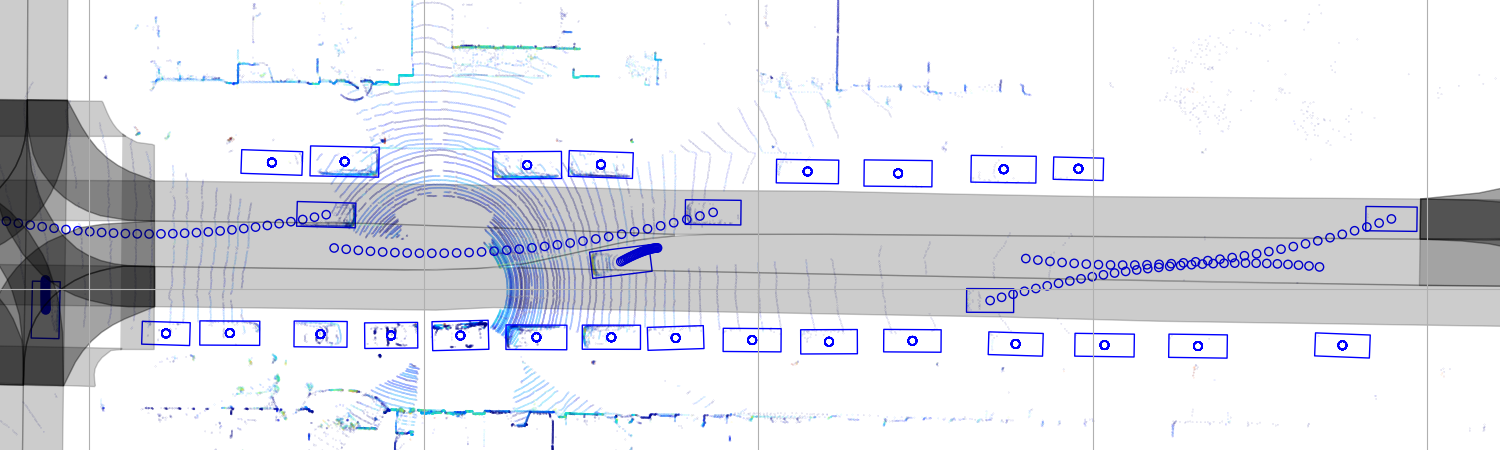}}
    \subfigure{\includegraphics[width=0.49\textwidth]{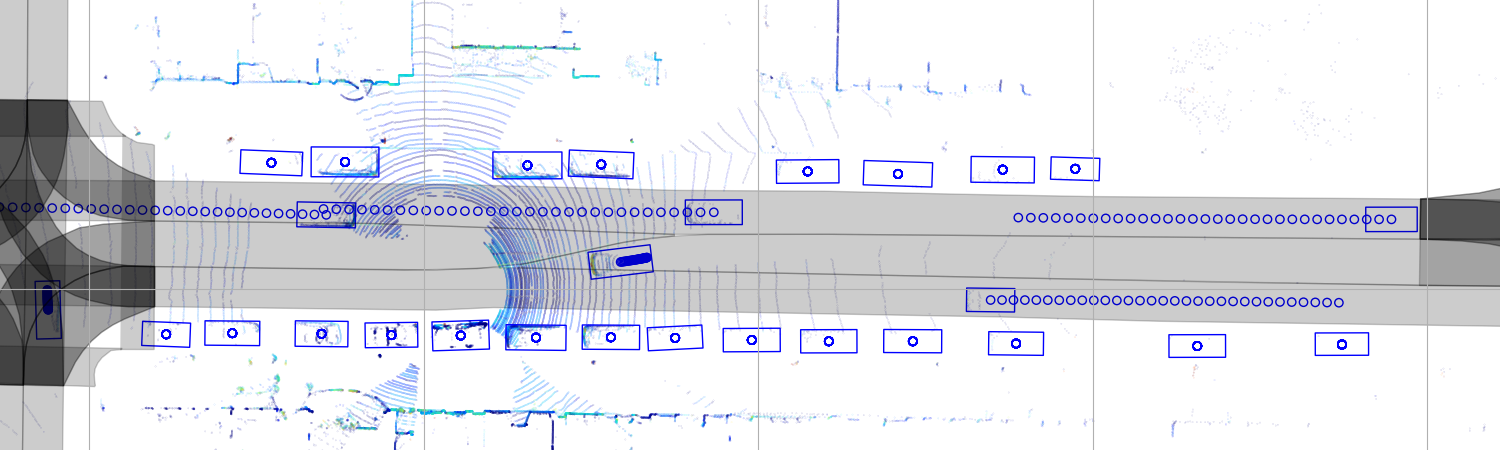}}
    \caption{Comparison of predicted trajectories without truncation (left) and with truncation (right) of the Gaussian raster; in the case of no truncation the minimization of the ellipse loss pushes the trajectories towards the middle of a drivable region
    }
    \label{fig:with_without_truncation}
    \vspace{-0.2cm}
\end{figure*}

\subsection{Quantitative results}

We trained \texttt{Ellipse-Loss} and the baselines three times, with means and standard deviations reported in Table~\ref{tab:pred-errors} where the best-performing methods are marked in bold.
We can see that the \texttt{Off-Road} method achieved better results than \texttt{MultiXNet}, where the additional penalty of off-road losses led to larger scene compliance without affecting prediction performance. 
In addition, the results show that \texttt{Ellipse-Loss} had significantly lower CtrORFP and BoxORFP ratios than either of the baseline methods.
In particular, it reduced both CtrORFP and BoxORFP by nearly a factor of $2$, while obtaining comparable $\ell_2$ error metrics. 

\subsection{Qualitative results}

In this section we present the case studies on three representative scenarios, shown in  Figure~\ref{fig:qualitative_results} which compares the predictions from \texttt{Ellipse-Loss} and \texttt{MultiXNet}.
In the first scenario the actor was merging into the lane from roadside parking.
We can see that the inferred trajectories from \texttt{MultiXNet} continued driving along the side of the road, while the prediction from \texttt{Ellipse-Loss} returned back to the road sooner, which was the ground-truth behavior as well.
In the second scenario the actor was making a U-turn in an intersection.
The left-hand figure shows that the prediction from \texttt{MultiXNet} went outside the road for the last few waypoints, while \texttt{Ellipse-Loss} trajectory was more scene-compliant and stayed inside the road.
In the third scenario the actor was making a left turn following a left-turning lane. 
We see that the prediction from \texttt{MultiXNet} did not follow the lane well, resulting in half of the bounding box going outside of its lane for the last few waypoints.
On the other hand, the prediction from \texttt{Ellipse-Loss} followed the lane much better, mirroring the ground-truth trajectory.
We can conclude that the novel box-aware loss yields more scene-compliant predictions, as it encourages the entire bounding box to stay inside the drivable region.

\subsection{Ablation study}
\label{sec:ablation_study}

We performed an ablation study with several variations of the ellipse loss to better characterize its effects.
In particular, we truncated the Gaussian raster at different Mahalanobis distances, with the results summarized in Table~\ref{tab:ablation_study_ellipse_sizes}.
For \texttt{2Md-ellipse} we truncated the loss at a Mahalanobis distance of $2$, which results in an ellipse that is $4 \times$ the size of the regular ellipse, denoted as \texttt{1Md-ellipse}.
Similarly, \texttt{0.5Md-ellipse} has $0.25 \times$ the ellipse size compared to \texttt{1Md-ellipse}.
For all metrics, we show both the absolute values and the relative deltas as compared to \texttt{1Md-ellipse}.
The results show that the off-road errors decreased but the ${\ell_2}$ errors increased as we increased the ellipse size.
This is expected, as due to increased ellipse loss the model focuses more on improving the scene compliance at the expense of less accurate predictions. 
On the flip side, the \texttt{0.5Md-ellipse} method had a lower ${\ell_2}$ error but significantly higher BoxORFP.
We can conclude that the \texttt{1Md-ellipse} model achieved the best balance between the two competing objectives.

Lastly, we also trained a \texttt{No-truncation} model without any truncation.
Expectedly, this model had lower \mbox{BoxORFP} but significantly higher ${\ell_2}$ errors than models with truncation.
As discussed previously, less truncation of the loss has the tendency to push the predicted trajectories closer to the center of drivable regions, affecting the model accuracy.
We visualize this characteristic in Figure~\ref{fig:with_without_truncation} on an example scene, showing results aligned with the toy experiments presented in Section \ref{sec:gradient analysis}.
To avoid such undesired behavior, it is important to truncate the Gaussian raster by the size of the actor's bounding box. 
This allows the ellipse loss to not over-penalize predicted bounding boxes that are already inside the drivable region, resulting in output trajectories that are both accurate and scene-compliant.


\section{Conclusion}

We considered the problem of motion prediction, a critical component of an autonomous system ensuring safe and efficient operations.
To improve the accuracy of output trajectories and ensure they obey the map constraints, we introduced a novel scene-compliant loss that takes into account the map and actor shape information.
This is done in a supervised manner, by projecting the trajectories into a top-down raster image using a recently proposed differentiable trajectory rasterizer, that helps align trajectory and map representations and directly penalize off-road predictions.
In addition, we showed how actor dimensions can be used to truncate the off-map loss, allowing for more precise loss computation.
To evaluate the proposed approach we used a large, real-world autonomous driving data, and extended a state-of-the-art model that performs joint detection and prediction with the proposed loss.
The results indicate that the method is significantly more realistic, while maintaining comparable prediction accuracy to the existing state-of-the-art approaches.
Moreover, detailed ablation and case studies provided important insights into the method's performance.

\balance

\newpage



\bibliographystyle{IEEEtran}
\bibliography{references}

\end{document}